\documentclass[letterpaper, 10 pt, conference]{./support/ieeeconf}
\usepackage{times}
\usepackage[pdftex]{graphicx}
\usepackage{subfigure}
\usepackage{amsmath,amssymb,amsopn,amstext,amsfonts}
\usepackage{cancel}
\usepackage[space]{cite}
\usepackage{pdfsync}
\usepackage{balance}
\usepackage{color}
\usepackage{mathtools}
\usepackage[ruled,vlined]{algorithm2e}
\usepackage{bm}

\usepackage{diagbox}
\usepackage{float}
\usepackage{epstopdf}
\usepackage{pifont}
\usepackage{fixltx2e}
\usepackage{amsmath}
\usepackage{multirow}
\usepackage{url}
\usepackage{svg}
\usepackage[linkcolor=black,citecolor=black,urlcolor=black,colorlinks=true]{hyperref}

\newcommand{\df}[1]{\mathrm{d}{#1}}

\newcommand{\norm}[1]{\Vert{#1}\Vert}

\graphicspath{{./figures/}}
\DeclareGraphicsExtensions{.png,.jpg,.eps,.pdf}
\IEEEoverridecommandlockouts
\title{\LARGE \bf Mapless-Planner: A Robust and Fast Planning Framework for Aggressive Autonomous Flight without Map Fusion}
\author{Jialin Ji*, Zhepei Wang*\thanks{*~Equal contributors.}, Yingjian Wang, Chao Xu and Fei Gao
\thanks{
  All authors are with the State Key Laboratory of Industrial Control Technology, and the Institute of Cyber-Systems and Control, Zhejiang University, Hangzhou, 310027, China.
  {\tt\small \{jlji, wangzhepei, yj\_wang, cxu, fgaoaa\}@zju.edu.cn}}%
}
\begin{document}

\maketitle
\thispagestyle{empty}
\pagestyle{empty}

\begin{abstract}
    Maintaining a map online is resource-consuming while a robust navigation system usually needs environment abstraction via a well-fused map. In this paper, we propose a mapless planner which directly conducts such abstraction on the unfused sensor data. A limited-memory data structure with a reliable proximity query algorithm is proposed for maintaining raw historical information. A sampling-based scheme is designed to extract the free-space skeleton. A smart waypoint selection strategy enables to generate high-quality trajectories within the resultant flight corridors. Our planner differs from other mapless ones in that it can abstract and exploit the environment information efficiently. The online replan consistency and success rate are both significantly improved against conventional mapless methods.
\end{abstract}
\section{Introduction}
\label{sec:introduction}

The autonomous navigation of quadrotors develops rapidly as it plays an important role in real-world applications. For a robust navigation framework, it is essential to build a well-fused map for back-end modules~\cite{tordesillas2019faster,fastPlanner,Oleynikova2020OpenMPF,zhou2020ego}. Maintaining a consistent map online requires reliable localization and sufficient computation power. Such requirements can make limited onboard resources inadequate for high-level missions other than navigation. It is attractive if robust navigation can also be achieved by a lightweight mapless planner.

Although there are many computationally efficient mapless planners~\cite{Lopez2017aggressive, florence2018nanomap, RAPPIDS}, most of them are trajectory-oriented. In other words, they focus on choosing a better feasible trajectory to accomplish collision avoidance. These planners do not extract the structure of the environment due to the lack of an informative map. Consequently, short-sighted planning occasionally occurs, thus easily leading the quadrotor to a dead end. Such planners are also sensitive to data noise, which can cause inconsistent replans. In comparison, the weaknesses are largely overcome in map-based planners such as~\cite{fastPlanner} and~\cite{gao2017gradient} because the map fusion already accomplishes a large part of the environment abstraction. To get over these weaknesses, it is important to incorporate such a mechanism into a mapless planner without fusing the sensor data.

\begin{figure}[t]
  \begin{center}
  \includegraphics[width=1.0\columnwidth]{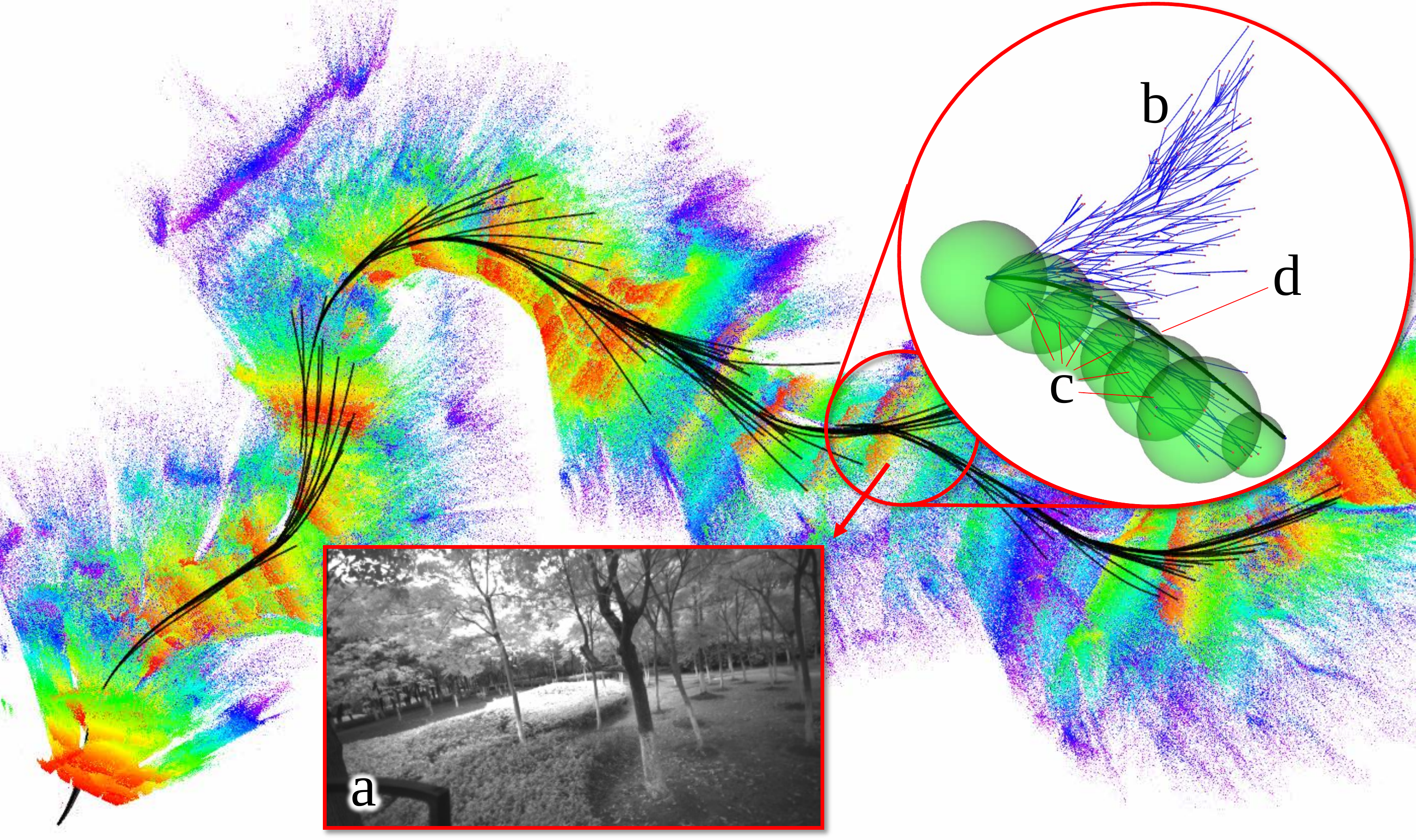}
  \end{center}
  \caption{
    \label{fig:introduction}
    A framework of mapless planning for autonomous quadrotors. (a) An onboard intensity image, (b) Forward spanning tree, (c) Ball-shaped flight corridor, (d) Trajectory generated via smart waypoint selection.
  }
\end{figure}

In this paper, we design a framework based on the above idea. Firstly, a limited-memory data structure is adopted for storing historical sensor information, following the basic idea of NanoMap. We further improve the reliability of query results and the compactness of un-fused data by redesigning a query algorithm. Secondly, a novel sampling-based scheme is utilized to accomplish the environment abstraction. It is able to extract the free-space skeleton and generate flight corridors simultaneously. Thirdly, a smart waypoint selection strategy is designed where our previous algorithm~\cite{wang2020alternating} serves as a sub-module. The scheme smartly inserts intermediate waypoint to suppress a trajectory into a corridor while maintaining its dynamic performance. It generates high-quality local trajectories in a computationally efficient way. Our framework abstracts and exploits the environment information in a relative cheap way. 

Summarizing our contributions as follows:
\begin{enumerate}
    \item We propose PicoMap, a limited-memory data structure to maintain historical sensor information. Adaptive downsampling and redesigned query interfaces are proposed for compact storage and accurate query.
    \item We propose Forward Spanning Tree, a sampling-based algorithm to extract free-space skeletons and generate safe flight corridors. All information in proximity query is exploited for the environment abstraction.
    \item We propose a lightweight yet effective trajectory generator which smartly selects waypoints to suppress a trajectory into a corridor. High-quality feasible trajectories are available with appropriate time allocation.
\end{enumerate}


\section{Related Work}
\label{sec:related_work}

\subsection{Data Structures for Mapless Planners}
For mapless planners, the sensor data is not fused into a prior map in an inertial frame such as the occupancy map. Instead, two commonly-used data structures in mapless methods are k-d Tree and Safe Flight Corridor (SFC) for environment abstraction. K-d tree enables proximity queries in surrounding environments. Directly building a k-d tree from recently received depth data is proposed in both~\cite{Lopez2017aggressive} and~\cite{florence2018nanomap} for mapless motion planning. On the contrary, SFC describes the free region in configuration space through combinations of geometric primitives. Bucki et al.~\cite{RAPPIDS} propose pyramid-shape free corridors directly generated from depth data such that motion primitives can be efficiently checked and conducted. Gao et al.~\cite{gao2019flying} propose a ball-shaped SFC directly generated from point clouds of LiDAR~\cite{Velodyne}. 

RAPPIDS~\cite{RAPPIDS} is a novel memoryless method that partitions the free space with pyramids given a single depth image. Although the generation of pyramids and collision checks can be quickly carried out, it is still prone to fail in complex environments, just like other memoryless methods due to their inherent limitation. NanoMap~\cite{florence2018nanomap} directly builds a k-d tree for each depth image. The history of k-d trees and their relative poses is utilized for proximity querying. It should be noted that the pose uncertainty is considered for queries in multiple k-d trees. However, its handling for occupied space and unknown space is not well designed, leading to inaccurate query results for corner cases. Moreover, the distance and direction information are not well utilized in the query results of k-d trees.

\subsection{Trajectory Generation for Mapless Planners}

Many mapless planners use motion primitives along with efficient feasibility checkers such as \cite{zhang2019maximum} and \cite{viswanathanefficient}. One main reason is that choosing feasible primitives only needs simple binary query results. Such kind of query has less restrictions on environment representation. Optimization-based planners enjoy high-quality trajectories while they require more abstraction on environments. SFC provides such abstraction so that convex formulations such as QP or QCQP are usually used in polynomial trajectory generation within SFC~\cite{tordesillas2019faster, gao2019flying, gao2020teach}. However, such methods cost more computation resources since the safety, dynamic limits, and smoothness should all be considered. For local trajectory generation, it is also not worthwhile to conduct trajectory optimization within an SFC with too many geometrical primitives. Our previous work~\cite{wang2020alternating} provides an efficient way to optimize a trajectory with prescribed waypoints while maintaining its feasibility. The way it handles waypoints and safety constraints lacks flexibility, which needs to be further improved with a smarter policy.

\section{Methodology}
\label{sec:methodology}

\subsection{PicoMap}

PicoMap gives further improvements over NanoMap\cite{florence2018nanomap}. It builds the 3D local data structure from each pair of the depth and intensity images. A proximity query algorithm is also designed for more reliable results.

\subsubsection{Building Algorithm} There  is no fusion procedure in the building stage. A finite-length list $\mathbf{L}_{maps}$ is maintained in memory. The list consists of a series of $\mathbf{kdTree}_i$ for all keyframes and transforms $\mathbf{TF}_i$ for all pairs of consecutive keyframes. There are two key ideas in this stage: 
\begin{itemize}
  \item $\mathbf{L}_{maps}$ is updated with a new keyframe only if a relative pose or relative time exceeds corresponding thresholds. This balances the coverage and the memory size of the limited historical sensor data.
  \item Adaptive down-sampling of the depth image is performed to build a compact k-d tree. More depth pixels are preserved if the corresponding region contains more textures in the intensity image.
\end{itemize}

\begin{algorithm}[t]
  \label{al2}
  \underline{\textbf{Function} safeRadiusQuery} ($\mathbf x_{query}$)

  \KwIn{~~$\mathbf x_{query}$, query point in world frame}
  \KwOut{$\mathbf r_{safe}$, safe radius \\ 
  ~~~~~~~~~~~$\mathbf x_{obstacle}$, nearest obstacle}
  \ForEach{$\mathbf{kdTree}_i$ and $\mathbf{TF}_i$ in $\mathbf L_{maps}$}{
    \If{IsInFov($\mathbf x_{query}$)}{
      $\Sigma_i \leftarrow$ EstimateStandardDeviation($\mathbf{TF}_i$) \\
      $\mathbf{d}_{edge} \leftarrow$ MinEdgeDistance$(\mathbf x_{query})$ \\
      $\mathbf{d}_{i}, \mathbf{x}_{i} \leftarrow$ NnSearch$(\mathbf x_{query})$ \\
      \If{$\mathbf{d}_{obstacle} < \mathbf{d}_{edge}$}{
        \Return{$(\mathbf{d}_{i}-\Sigma_i), \mathbf x_{obstacle}$}
      } \ElseIf{$\mathbf{d}_{i} - \Sigma_i < r_{safe}$}{
        $\mathbf r_{safe} \leftarrow \mathbf{d}_{i} - \Sigma_i$ \\
        $\mathbf x_{obstacle} \leftarrow \mathbf x_{i}$
      }
    }
  }
  \If{never seen in FOV}{
    $\mathbf r_{safe}, \mathbf{x}_{obstacle} \leftarrow$ NnSearch$(\mathbf x_{query})$
  }
  \Return{$\mathbf r_{safe}, \mathbf{x}_{obstacle}$}  
\caption{PicoMap query algorithm}
\end{algorithm}

\begin{figure}[t]
  \centering
      \subfigure[\label{fig:f01}]
  {\includegraphics[height=0.47\columnwidth]{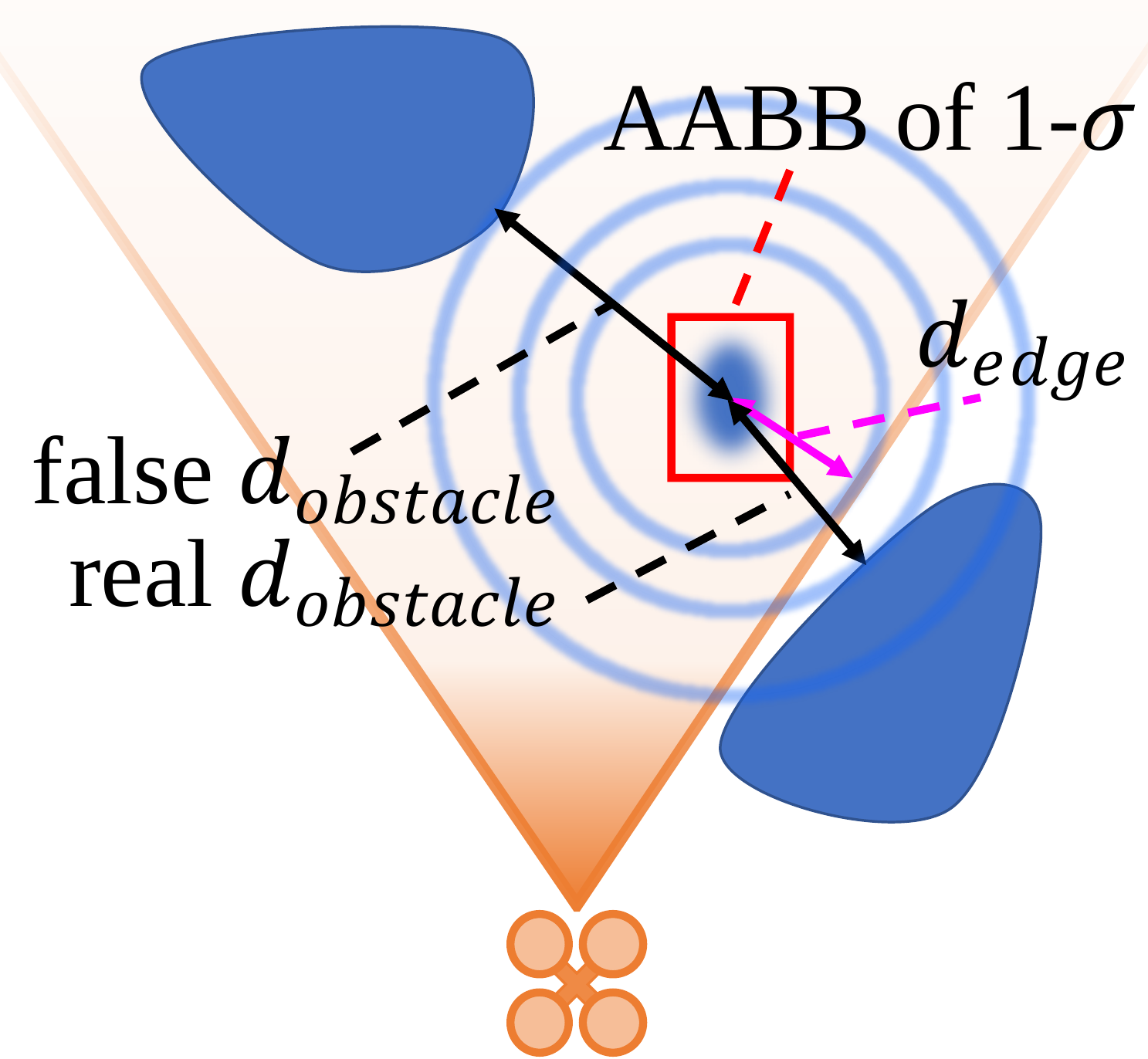}}
      \subfigure[\label{fig:f00}]
  {\includegraphics[height=0.47\columnwidth]{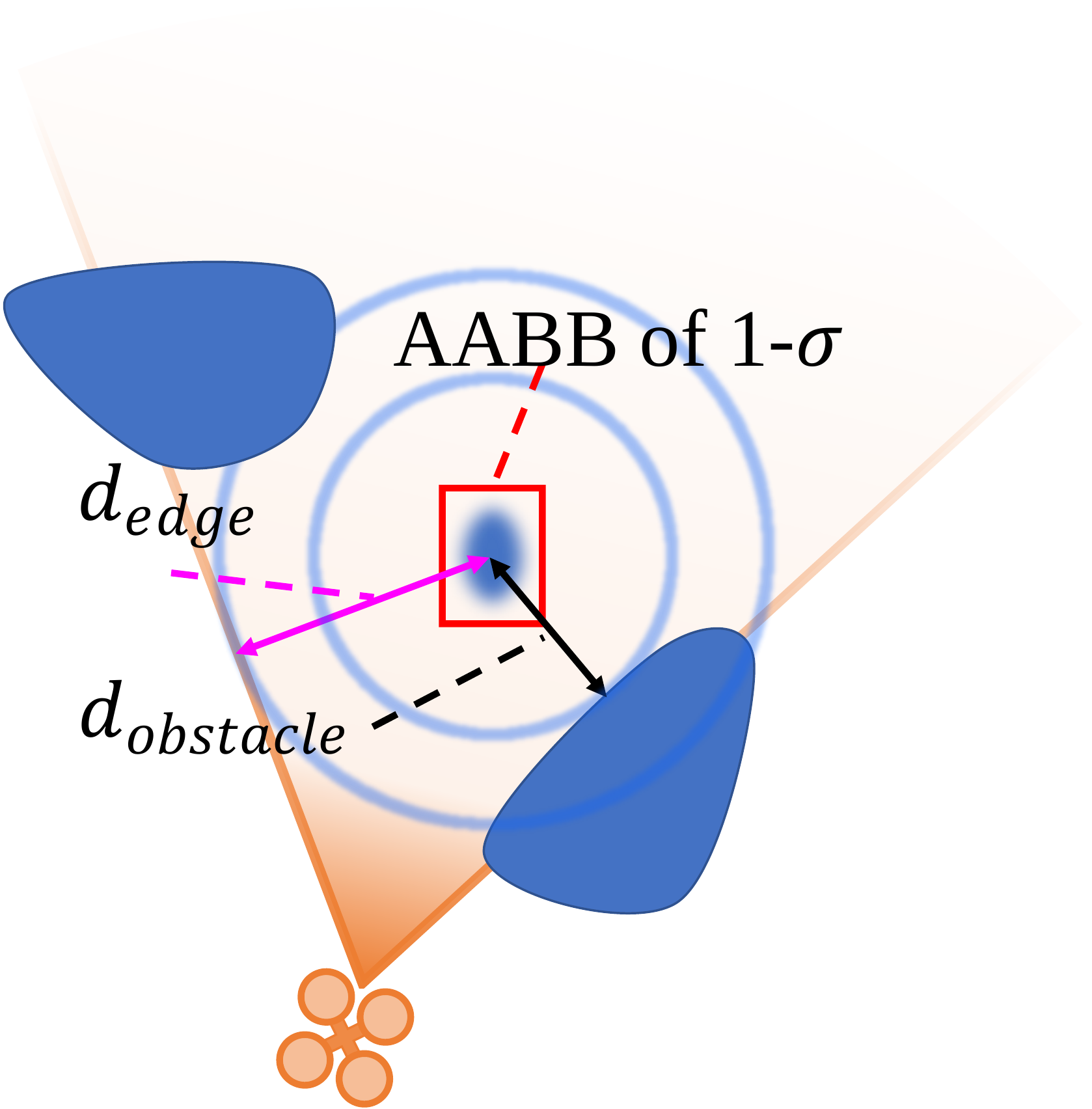}}
  \caption{The safe radius $r_{safe}$ of the query point at the edge of FOV may be overestimated by NanoMap. (a) $\mathbf d_{obstacle} > \mathbf d_{edge} \nRightarrow r_{safe} = \mathbf d_{obstacle} $, (b) $\mathbf d_{obstacle} < \mathbf d_{edge} \Rightarrow r_{safe} = \mathbf d_{obstacle}$}
  \vspace{-1cm}
\end{figure}

\subsubsection{Querying Algorithm}
Following NanoMap~\cite{florence2018nanomap}, we also iteratively check whether the query point is in the FOV (IsInFov()) and estimate uncertainty of the query (EstimateStandardDeviation()) until we find it, as is shown in Algorithm \ref{al2}. Then, the nearest obstacle $x_{obstacle}$ is queried through k-d tree (NnSearch()). 

For a query point near the boundary of FOV, NanoMap\cite{florence2018nanomap} checks the AABB of the 1-$\sigma$ of the query point distribution. However, one circumstance in Fig. \ref{fig:f01} may occur that the AABB lies in FOV but $\mathbf{d}_{obstacle} > \mathbf{d}_{edge}$. Then, $\mathbf{d}_{obstacle}$ is overestimated considering the unseen $\mathbf{x}_{obstacle}$. Under such circumstance we should continue iteratively checking whether there is a more appropriate view ($\mathbf{d}_{obstacle} < \mathbf{d}_{edge}$) in $\mathbf{L}_{maps}$ as Fig. \ref{fig:f00} for instance. 
If the query point is never seen in any FOV, we regard $x_{obstacle}$ queried in the first frame as the nearest obstacle optimistically, avoiding problems from being occluded by very close objects.

\subsection{Forward Spanning Tree}
Due to the limited sensing range, a quadrotor seldom turns around in a single local replan if its intermediate goal is appropriately selected. We make such an assumption that a local trajectory is always inside-out, i.e., flying away from the current position. According this assumption, we propose an efficient sampling-based method to extract the skeleton of free spaces using SFCs. 

\begin{figure}[t]
  \centering
  \subfigure[\label{fig:f1}]
  {\includegraphics[height=0.50\columnwidth]{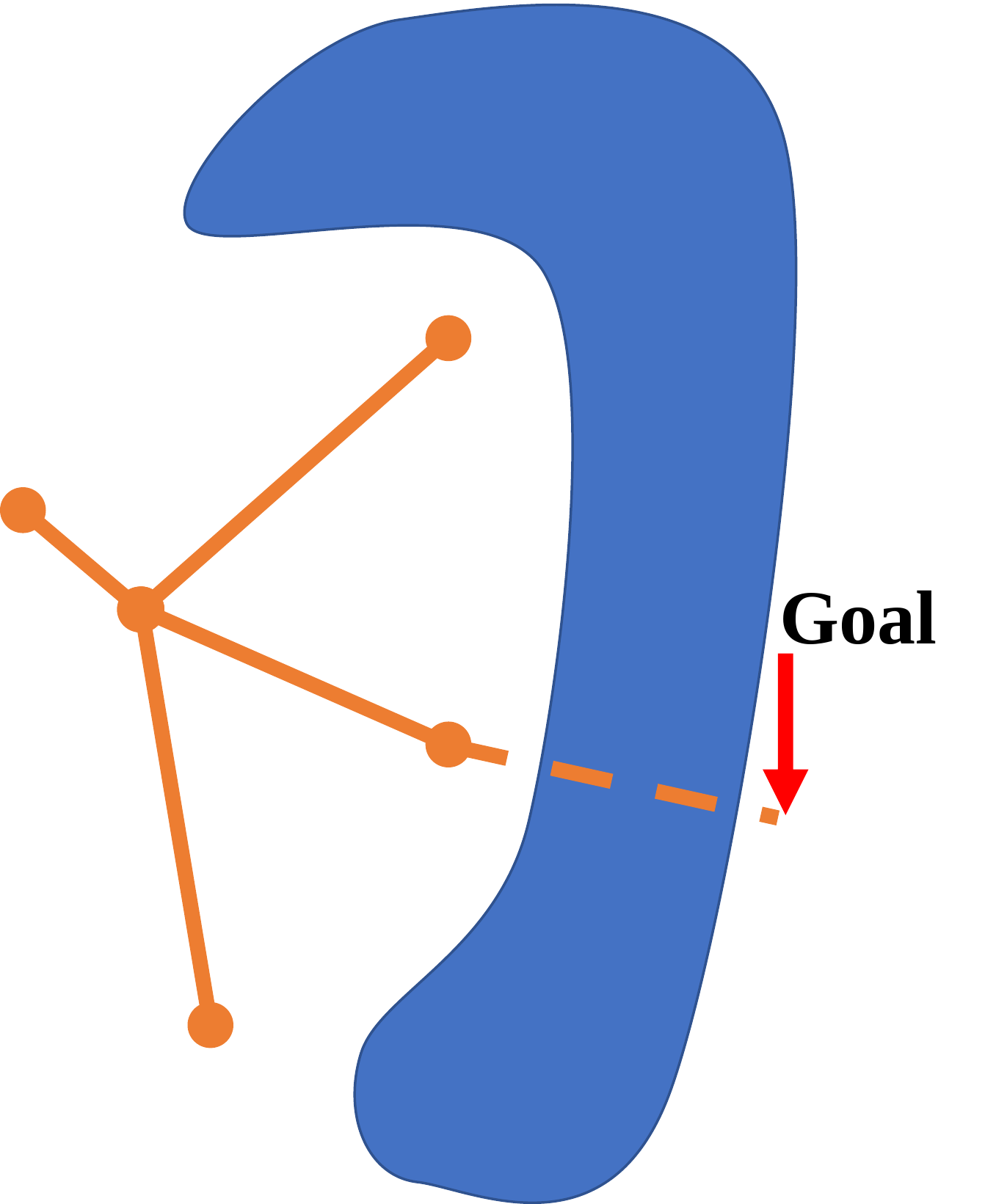}}
  \subfigure[\label{fig:f2}]
  {\includegraphics[height=0.50\columnwidth]{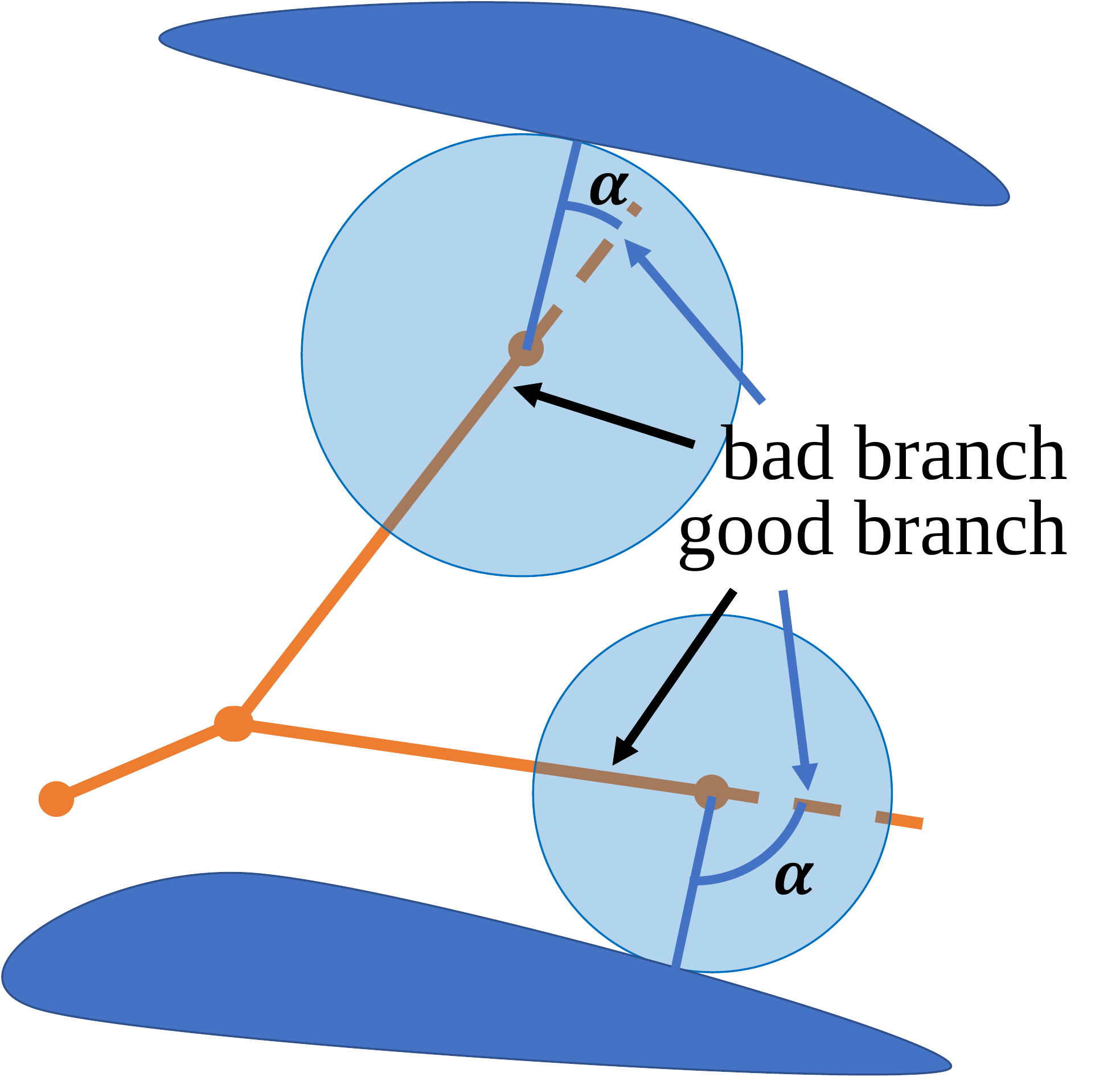}}
  \caption{(a) A branch falling into a dead end. (b) Score the leaf nodes according to the angle $\alpha$ between spanning direction and the nearest obstacle.}
\end{figure}

\begin{figure}[t]
  \centering
      \subfigure[\label{fig:f3}]
  {\includegraphics[height=0.45\columnwidth]{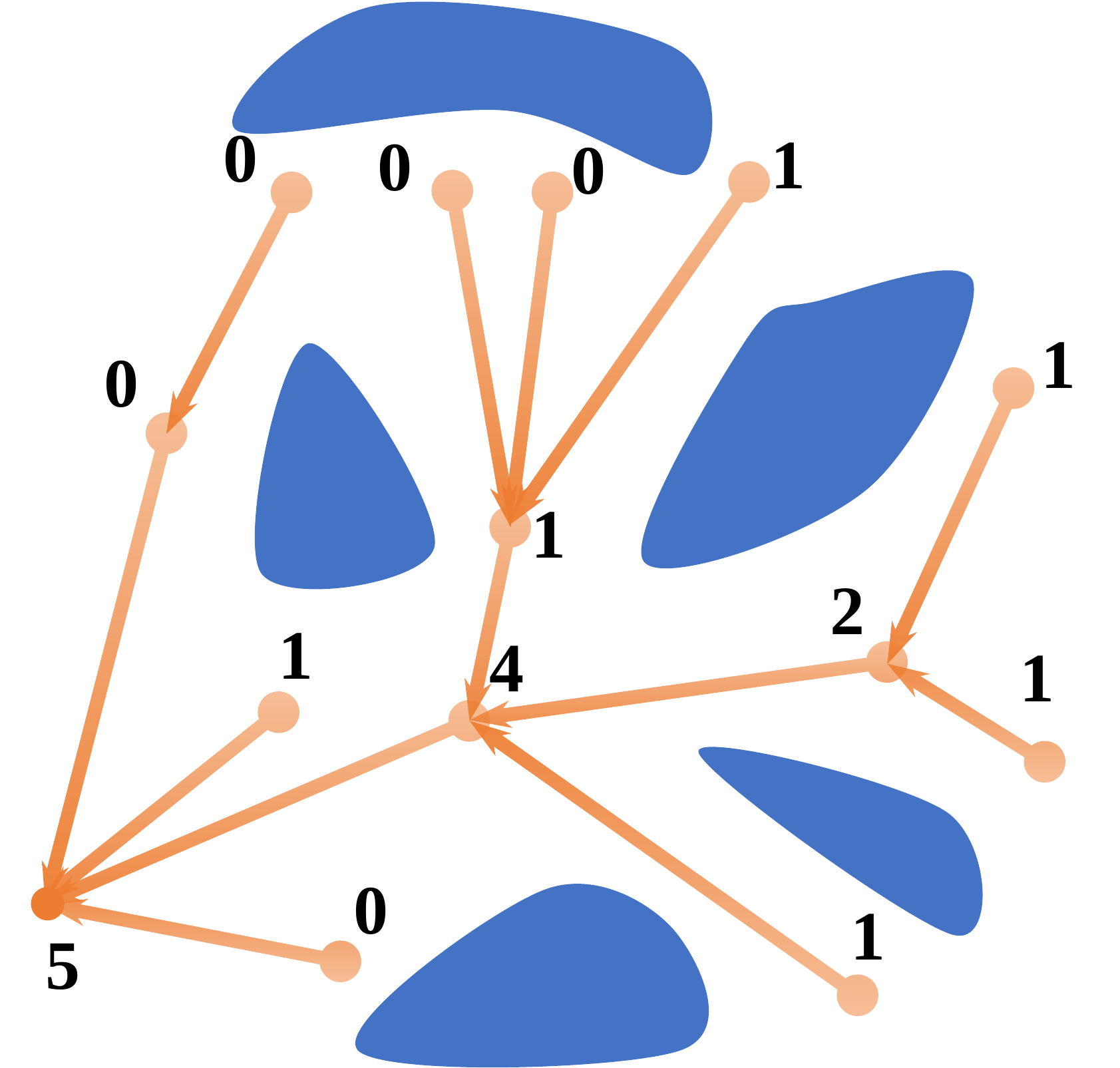}}
      \subfigure[\label{fig:f4}]
  {\includegraphics[height=0.45\columnwidth]{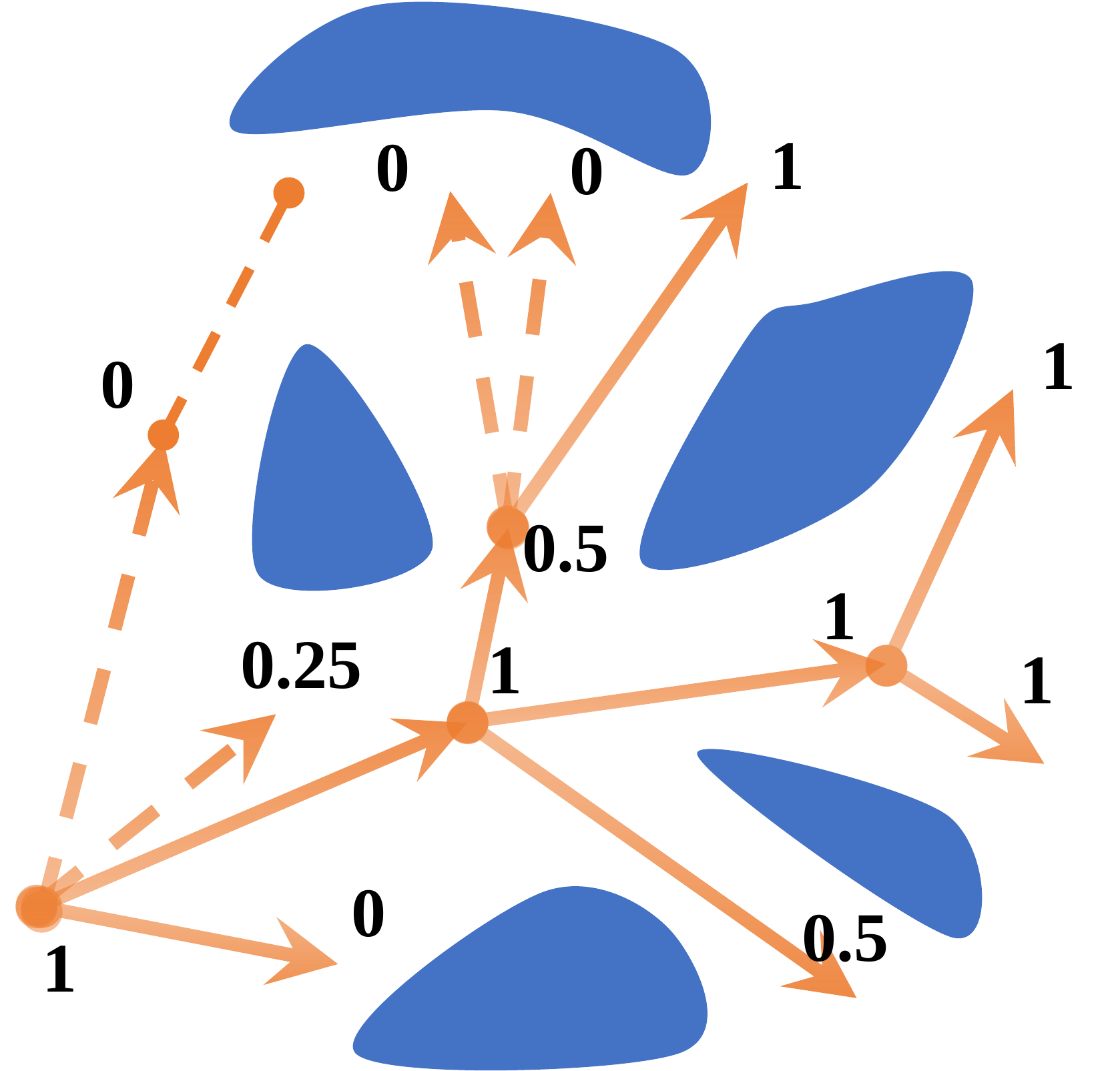}}
  \caption{Score (a) and prune (b) the forward spanning tree.}
  \vspace{-0.5cm}
\end{figure}

\subsubsection{Building Algorithm} A batch of free samples is first generated using PicoMap. According to our assumption, all samples are firstly sorted by the distance from the root node (Sort() in Algorithm \ref{al3}) and then inserted to the forward spanning tree (FST) one by one. The cost of each inserted node is calculated by the sum of the cost of its parent and the distance of the node from its parent. For each inserted data, in order to choose a reasonable parent node, a dynamic k-d tree~\cite{bentley1980decomposable} is maintained to provide the Knn() function. Among the $n$ nearest neighbors which are able to be connected to the inserted data without collision, the node whose cost is lowest ($N_p$) should be chosen as the parent of the inserted data. It should be noted that the time-consuming rewire operation is avoided by using Sort() based on our assumption.

\begin{algorithm}[t]
  \label{al3}
  \KwIn{~~$\mathbf{P}_{root}$, position of root node\\
    ~~~~~~~~~~~$\mathbf S_{sample}$, batch sampling data}
  \KwOut{FST, forward spanning tree}
  insert $\mathbf{P}_{root}$ as the root node of FST \\
  $\mathbf S_{sample} \leftarrow$ Sort($\mathbf S_{sample}$) \\
  d-KdTree $\leftarrow$ BuildDynamicKdTree(FST) \\
  \ForEach{$\mathbf P_{i}$ in $\mathbf S_{sample}$}{
    $\mathbf{dist} \leftarrow $ safeRadiusQuery ($\mathbf P_{i}$) \\
    \If{$\mathbf{dist} < \mathbf{r}_{inflate}$} {
      $\mathbf{S}_{knn} \leftarrow$ Knn($\mathbf P_{i}$) \\
      \ForEach{${N_i}$ in $\mathbf{S}_{knn}$}{
        $Cost_i \leftarrow N_i$.cost $+||{N_i}$ - $P_{i}||_2$
      }
      $p \leftarrow \arg\min(Cost_i)$ \\
      append child $\mathbf P_{i}$ to $N_p$ \\
      $\mathbf P_{i}$.cost $\leftarrow N_p$.cost $+||{N_p}$ - $P_{i}||_2$ \\
      rebuild d-KdTree
    }
  }
  \Begin(node $N_i$ for post-order traversal of FST){
    \eIf{$N_i$ is leaf node}{
      $N_i$.score $\leftarrow$ Score($N_i$)
    } {
      $N_i$.parent.score $\leftarrow$ $N_i$.parent.score + $N_i$.score
    }
  }
  \Begin(node $N_i$ for breadth-first traversal of FST){
    $M \leftarrow \max$ score of siblings of $N_i$ \\
    $w \leftarrow$ $N_i$.score / $M$ \\
    \If{$w < prune factor$} {
      cut off branch of $N_i$
    }
  }
  \KwRet{FST}
\caption{Build and prune forward spanning tree}
\end{algorithm}

\subsubsection{Pruning Algorithm}
As is shown in Fig.\ref{fig:f1}, the resultant tree contains redundant information on the environment, from which we wish to extract useful SFCs. We propose a two-stage scheme. In the scoring stage, we first score each leaf node based on its angle $\alpha$ between the spanning direction and the nearest obstacle. If an $\alpha$ is smaller than a prescribed threshold, the corresponding leaf is spanning forward to the obstacle. We set its score to 0. Otherwise, the leaf in Fig.\ref{fig:f2}, whose $\alpha$ is larger than the threshold, is spanning along the skeleton of the fee space. We set the score of such a leaf to 1. The score of any non-leaf node is defined as the sum of its children's. Following this rule, scoring the entire tree can be carried out by post-order traversal of FST as is given in Fig.\ref{fig:f3}. In the pruning stage, the pruning weight $w$ for each node is defined by its score divided by the max score of its siblings $M$. A branch is pruned from the tree if the weight of its root is lower than a prescribed prune factor. The pruning procedure can be carried out by the breadth-first traversal of FST, as is shown in Fig.\ref{fig:f4}. As a result, the remaining branches are all spanning forward to free space.

\subsubsection{Ball-shaped SFC Generation}
\begin{figure}[t]
  \centering
      \subfigure[\label{fig:f5}]
  {\includegraphics[height=0.39\columnwidth]{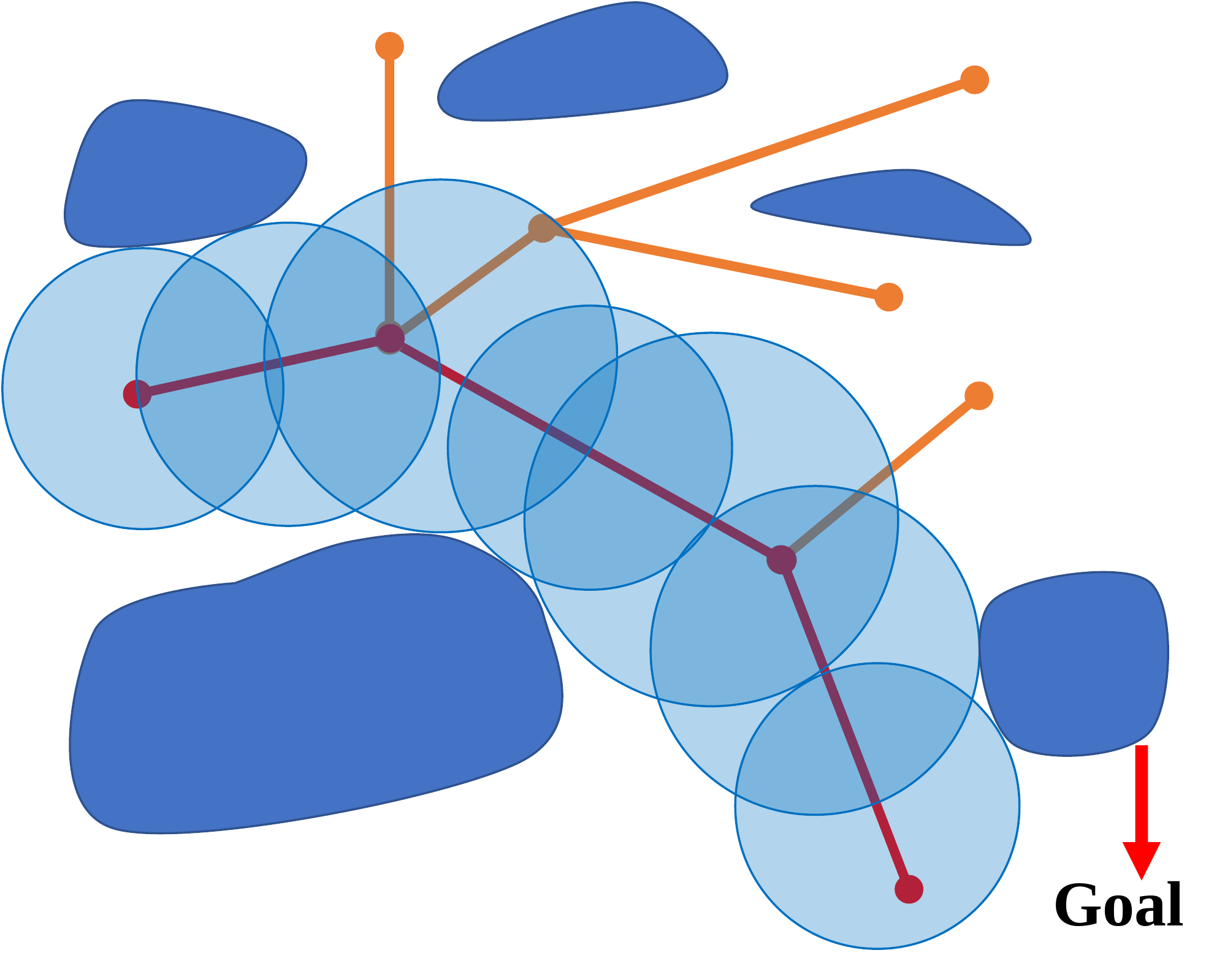}}
      \subfigure[\label{fig:f7}]
  {\includegraphics[height=0.39\columnwidth]{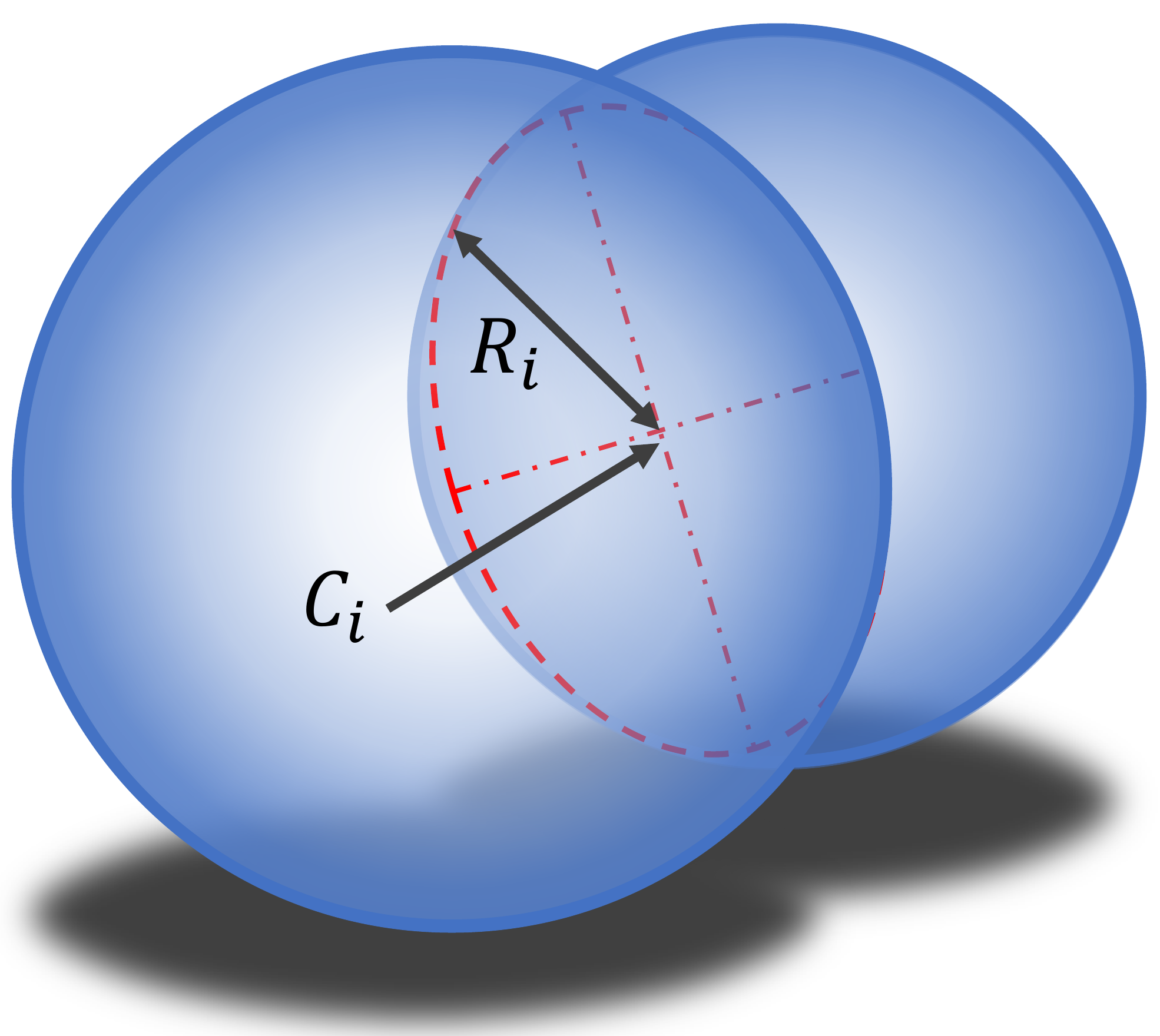}}
  \caption{(a) Generation of the ball-shaped SFC, (b) Intersection of the surfaces of adjacent two balls}
  \vspace{-1cm}
\end{figure}

Provided with a goal, the nearest leaf node along its path from the root is generated from the pruned FST, which is the red line in Fig.\ref{fig:f5}. Then, a ball-shaped SFC is generated along this path. Any ball is centered at the intersection of its predecessor and the path. All radii of these balls are obtained through safeRadiusQuery(). 

\subsubsection{Dynamic Resampling}
In order to improve the consistency of two consecutive replans, we sample from a normalized weighted sum of Gaussian distributions generated from previous results. The weight, center, and covariance of each Gaussian distribution are determined by the volume, position, and radius of the free ball of a point in the previous frame. Such kind of dynamic resampling also improves the sampling rate.

\subsection{Trajectory Generation}
We denote by $\mathcal{F}$ the resultant SFC which consists of $M$ sequentially connected balls $\mathcal{B}_i$, i.e., $\mathcal{F}=\bigcup_{i=1}^{M}\mathcal{B}_i$. We wish to generate an energy-time optimal trajectory $p(t):[0,T]\mapsto\mathbb{R}^3$ confined by $\mathcal{F}$:
\begin{align}
\min_{p(t)}&~{\int_{0}^{T}{\norm{p^{(3)}(t)}^2\df{t}}+\rho T},\\
s.t.~&~p(0)=p_o,~p^{(1)}(0)=v_o,~p^{(2)}(0)=a_o,\\
&~p(T)=p_f,~p^{(1)}(T)=v_f,~p^{(2)}(T)=a_f,\\
&~\norm{p^{(1)}(t)}\leq v_m,~\norm{p^{(2)}(t)}\leq a_m,~\forall t\in[0,T],\\
&~p(t)\in\mathcal{F},~\forall t\in[0,T],
\end{align}
where $p(t)$ is a polynomial spline over $[0,T]$, $\rho$ the weight for time regularization, $v_m$ and $a_m$ the dynamic limits. Solving such a problem is quite difficult and not worthwhile for local trajectories. Therefore, we design a heuristic procedure to approximate the optimal trajectory of the above problem.

To retain both efficiency and quality, we propose a smart waypoint selection strategy for trajectory generation within an SFC. The alternating minimization scheme in our previous work~\cite{wang2020alternating} is served as a sub-module of this scheme. Once all waypoints are given, this sub-module can generate a $p(t)$ with appropriate time allocation and guaranteed feasibility. Our waypoint selection strategy borrows an idea from the Douglas-Peucker algorithm~\cite{douglas1973algorithms}. Intuitively, the maximal safety violation of a trajectory is taken as the highest priority, thus it is suppressed by a newly-added waypoint.

\begin{figure}[t]
  \begin{center}
  \includegraphics[width=1\columnwidth]{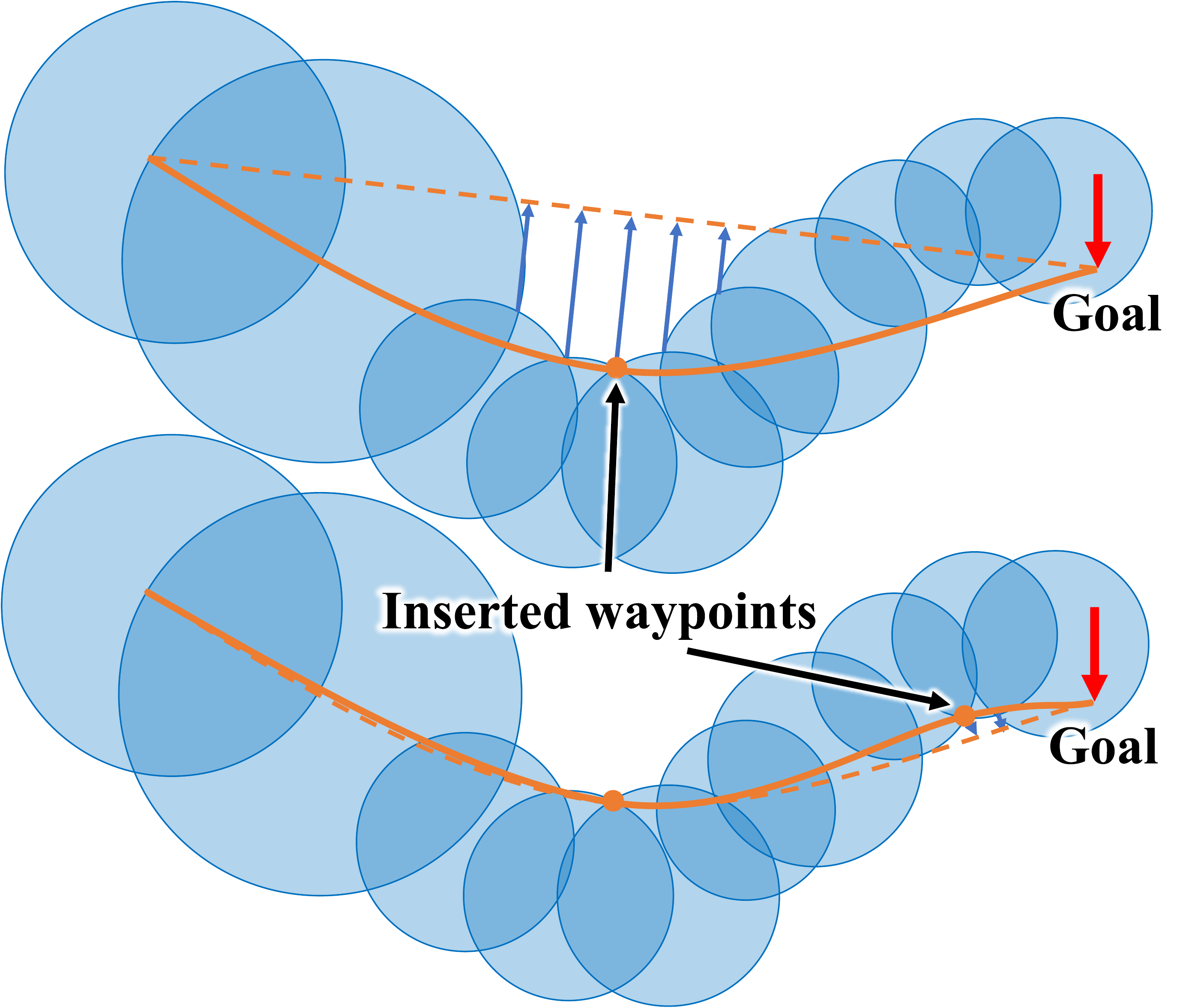}
  \end{center}
  \caption{
    \label{fig:f6}
    A trajectory generated based on our waypoint selection strategy.
  }
  \vspace{-1cm}
\end{figure}

Firstly, each circle is generated by intersecting a pair of two adjacent balls. As is shown in Fig.\ref{fig:f7}, we denote by  $C_i$ its center and by $R_i$ its radius, as is shown in Fig.\ref{fig:f7}. The initial trajectory is a one-piece 5-order polynomial calculated from the boundary condition. As is shown in Fig.\ref{fig:f6}, the initial trajectory should be a straight line in a rest-to-rest case. The following iterations are conducted for safety constraints:
\begin{itemize}
  \item \textit{Step 1}: Calculate each distance $d_i$ from the $i$-th circle to the trajectory. If the trajectory crosses the circle, set the $d_i$ to zero.
  \item \textit{Step 2}: Find the circle with the maximal $d_i$. Add the closest point on the circle as a new waypoint of the trajectory.
  \item \textit{Step 3}: For the given boundary condition and waypoints, use the algorithm proposed in ~\cite{wang2020alternating} to calculate the whole trajectory. 
  \item \textit{Step 4}: Repeat \textit{Step 2} and \textit{Step 3} if there still exists a positive $d_i$. Otherwise, the whole trajectory is already within the SFC.
\end{itemize}

During the flight, a newly generated trajectory is committed to the tracking controller if its distance to goal is smaller than the currently tracked one.
\section{Experiments and Benchmarks}
\label{sec:experiments}

\begin{figure*}[t]
  \begin{center}
    \subfigure[\label{fig:simulation1}]{
      \includegraphics[height=0.65\columnwidth]{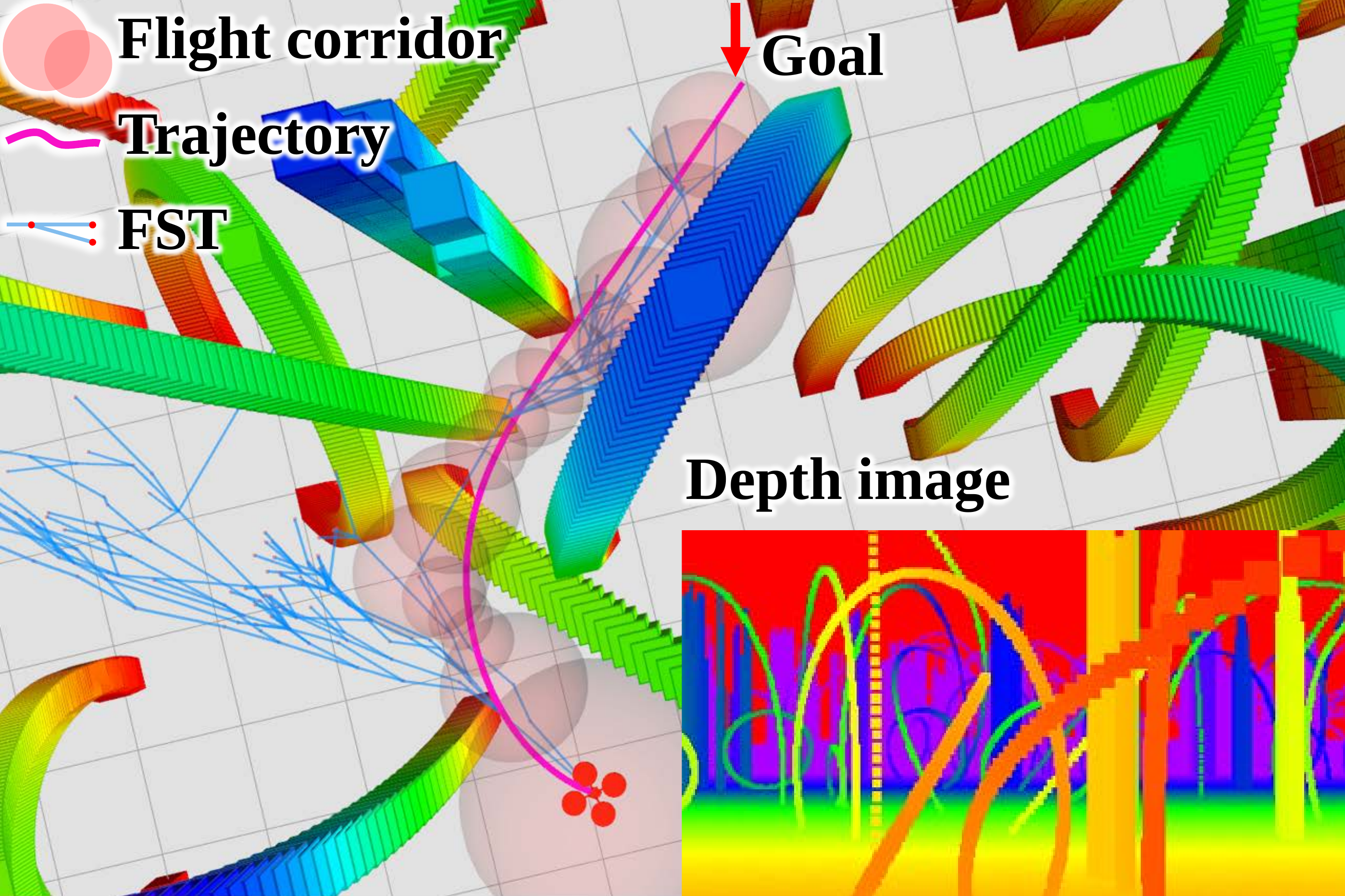}
    }
    \subfigure[\label{fig:simulation2}]{
      \includegraphics[height=0.65\columnwidth]{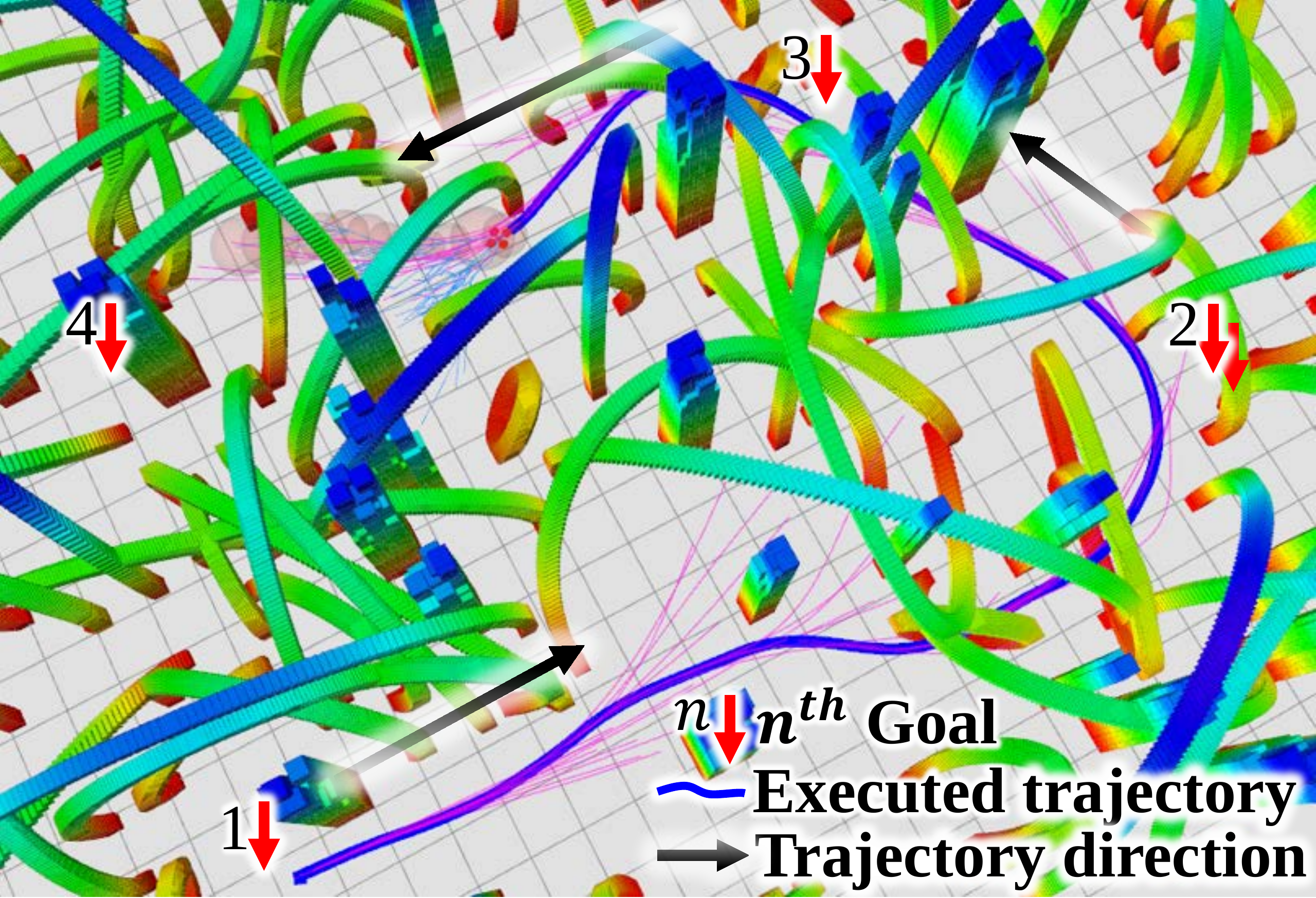}
    }
  \end{center}
  \caption{
    (a) Simulation in a cluttered environment. (b) The drone generates a consistent trajectory when the goal changes while flying.
  }
\end{figure*}

\begin{figure*}[t]
  \begin{center}
  \includegraphics[width=2\columnwidth]{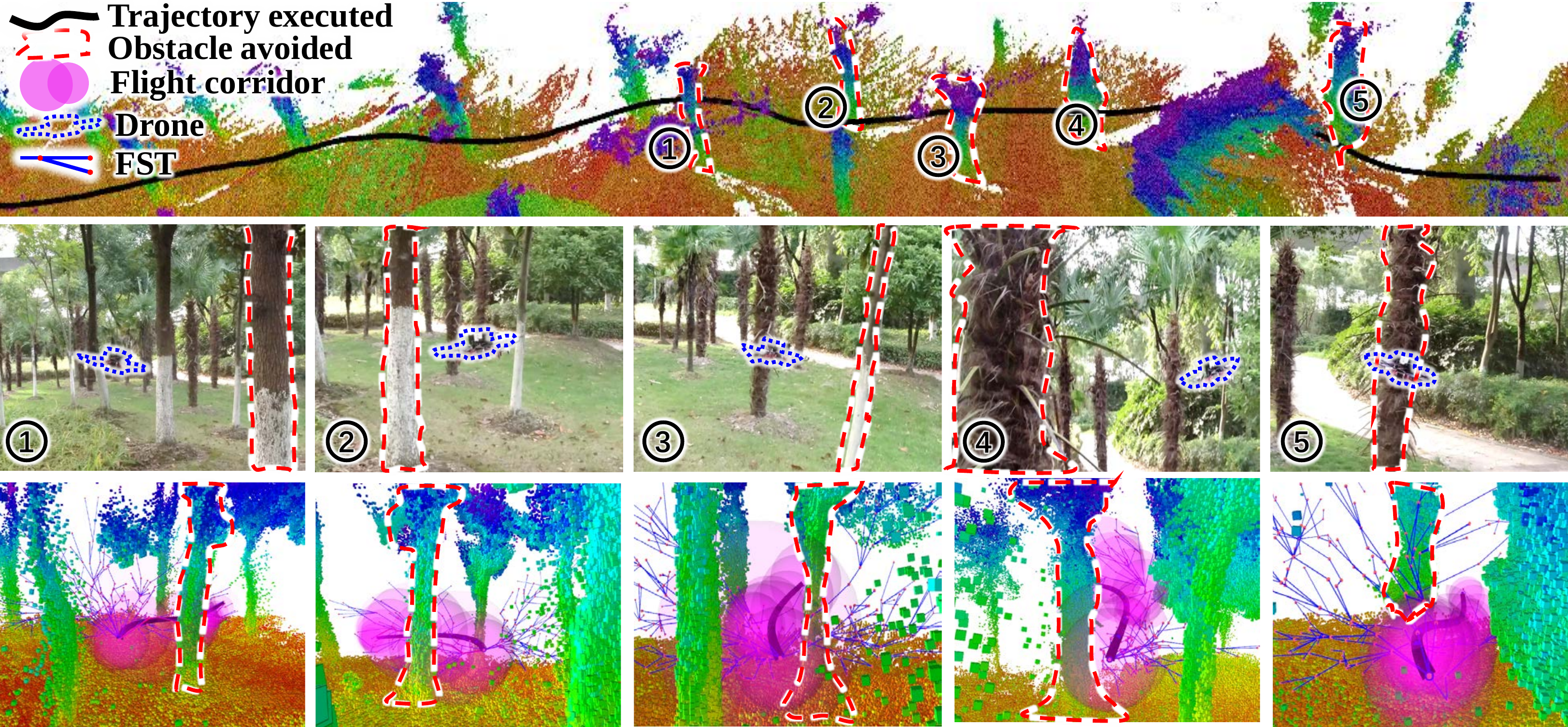}
  \end{center}
  \caption{
    \label{fig:outdoor}
    An outdoor experiment in a dense forest. The drone avoid the obstacles one by one rapidly with expected speed of 3m/s. 
  }
\end{figure*}

\subsection{Simulation in Cluttered Environment}

To validate the robustness of the proposed method, we conduct several simulations in randomly generated cluttered environments. The environments consist of $300$ pillar-shaped obstacles and $120$ ring-shaped obstacles in a $30m\times30m\times5m$ field. In such scenes, RAPPIDS\cite{RAPPIDS} cannot generate pyramids that describe the free space exactly. In comparison, our pruned FST precisely captures the skeleton of the free region as is shown in Fig.~\ref{fig:simulation1}. Based on the preferred path, an SFC is generated for subsequent trajectory optimization. Each ball in the SFC expanses as large as possible, showing the accurate query result of our PicoMap. In Fig.~\ref{fig:simulation2}, consistent trajectories are generated when the goal is changed online.

\subsection{Real-world Experiments}

To validate the practical performance of our method, we deploy it on a compact quadrotor platform whose configuration is detailed in \cite{ji2020cmpcc}. The real-world experiment is conducted in a dense forest shown in Fig.~\ref{fig:outdoor}. A filter-based visual-inertial odometery~\cite{sun2018robust} is utilized as the feedback for the trajectory tracking controller. The drone avoids the obstacles one by one rapidly with an expected speed of $3m/s$, showing its aggressiveness and robustness. In Fig.~\ref{fig:outdoor}, the pruned FST bypass all the trees, guiding the SFC into large free regions. The generated trajectory is also confined by the SFC with a few of waypoints inserted, which validates the smartness of our waypoint selection strategy.

\subsection{Modular Tests of the Framework}

\begin{figure}[ht]
  \begin{center}
    \includegraphics[height=0.4\columnwidth]{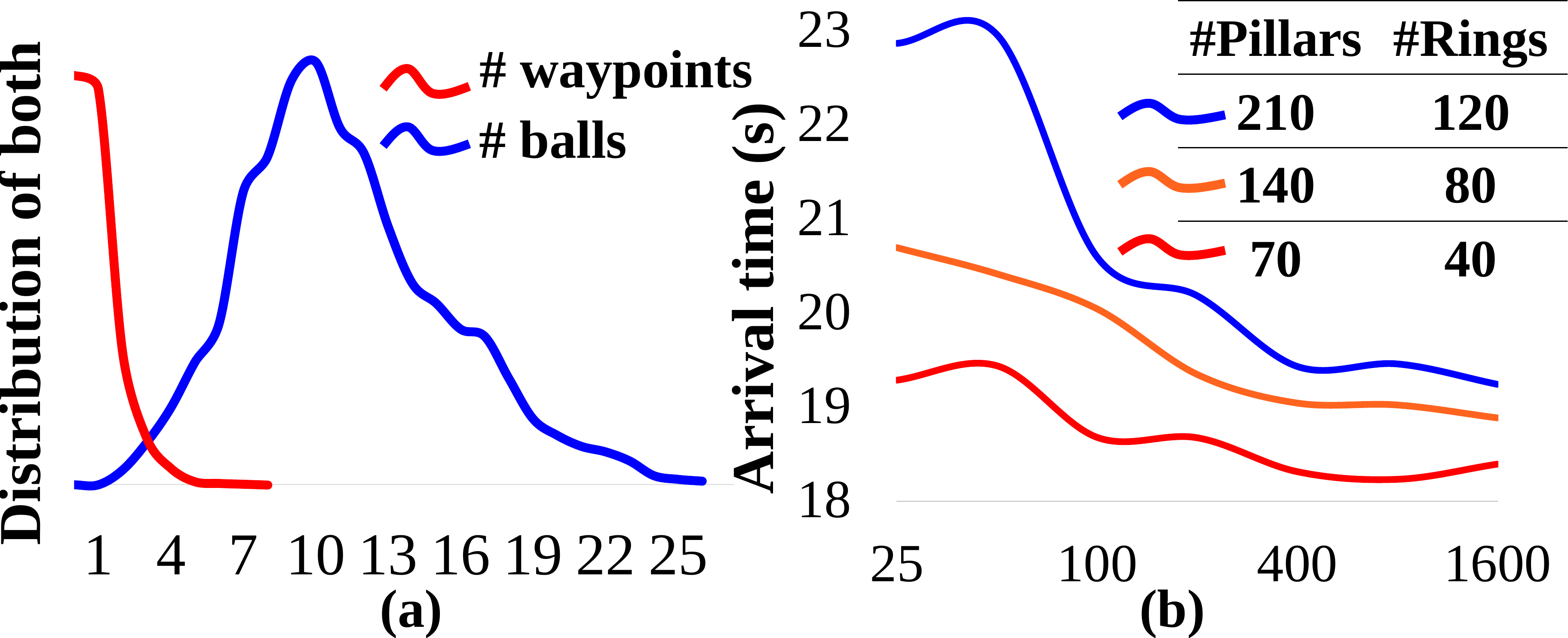}
  \end{center}
  \caption{
    \label{fig:moretests}
    (a) Distribution of the number of balls and waypoints inserted. (b) Arrival time of different sampling data for FST in different environments.
  }
\end{figure}

In order to quantify the performance of each part of the framework, we set three different tests: 

(1) For the generation of SFCs and trajectories, we record the distribution of the numbers of both balls and waypoints inserted to generate a feasible trajectory for $1272$ tests, shown in Fig. \ref{fig:moretests}(a). Most of the flight corridors are of size $10$ but only $7\%$ of the number of  waypoints inserted to the trajectories are larger than $3$. Moreover, $80\%$ of the trajectories are feasible by inserting no more than one additional waypoint. These statistics indicate the smartness of our trajectory generation method. 

(2) For the FST, we test the performance of different sampling numbers in random $20m\times40m\times5m$ environments for $100$ times, as is shown in Fig. \ref{fig:moretests}(b). The arrival time decreases as we increase the size of batch sampling. When sampling size exceeds $200$, the shortening of arrival time becomes insensitive. Thus the sampling size of $200$ samples suffices for planning in such kinds of environments. 

(3) We also record the average computation time of building a k-d tree from a depth image, obtaining a pruned FST and generating a trajectory by averaging over $2259$ times on Intel i7-6700 CPU. Specifically, the resolution of depth and intensity image is $640\times480$. The sampling number of FST is $200$. The average computation time of each part is $3.24ms$, $0.98ms$ and $0.26ms$, respectively. It is worth noting that over half of the time is spent on image pixel accessing.

\subsection{Benchmark}

\begin{figure}[ht]
  \begin{center}
  \includegraphics[width=1\columnwidth]{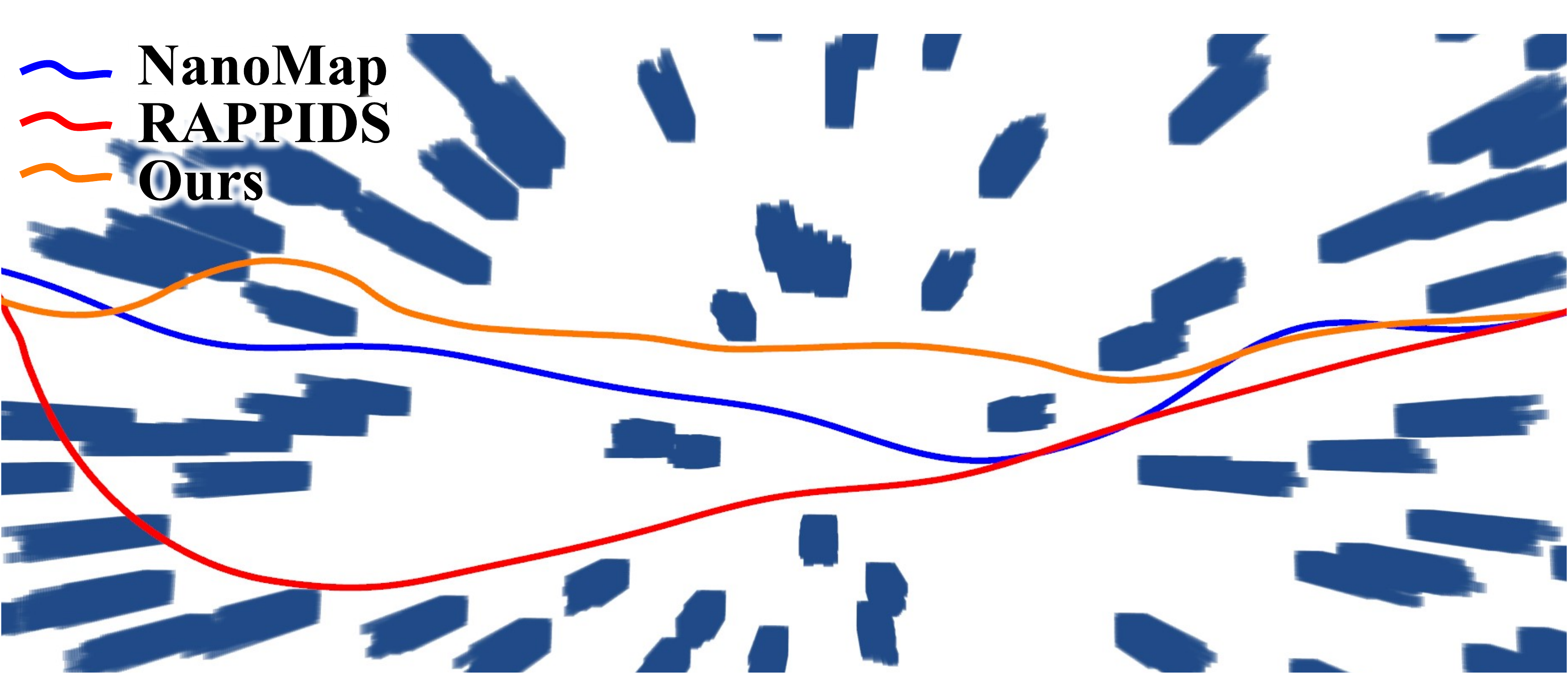}
  \end{center}
  \caption{
    \label{fig:benchmark1}
    A random 20m$\times$40m enviroment for benchmark that consists of randomly placed pillars and circular rings. Compared to that RAPPIDS and NanoMap can only choose some open space to pass through, Ours can get to goal faster by flying through some narrow spaces aggressively.
  }
\end{figure}

We evaluate the overall planner performance of our algorithm in terms of safety and aggressiveness. The benchmark is conducted using our method and some other state-of-the-art mapless approaches, i.e., RAPPIDS and NanoMap, as is shown in Fig. \ref{fig:benchmark1}.

Specifically, the onboard depth sensor provides depth and intensity images at $640\times480$ resolution, which is the same as is used in real-world experiments. The field of view (FOV) is set to $78$ degrees horizontally and $64$ degrees vertically. The real-time poses and images of the camera are provided by the simulator. Both sensors are simulated at $30$Hz. The experiments are repeated for $100$ times in randomly generated environments under different difficulty. The difficulty is quantified by the expected maximum flight velocity $\mathbf{V}_{max}$ and the obstacle density $\mathbf{I}_{obs}$. These two parameters are chosen as $\mathbf{V}_{max} = {\lbrace2, 3, 4\rbrace}~m/s$ and $\mathbf{I}_{obs}= {\lbrace 30, 60, 90\rbrace}$ obstacles, for a total of $3\times3\times100\times3 = 2700$ simulation trials. The distance between the initial position and goal is $18m$. In each case, a timer is started when the quadrotor takes off and is stopped when a collision occurs or the goal is reached. One flight mission is considered to be successful only if the target is reached in a limited time. 

\begin{figure}[ht]
  \centering
  \includegraphics[width=1.0\columnwidth]{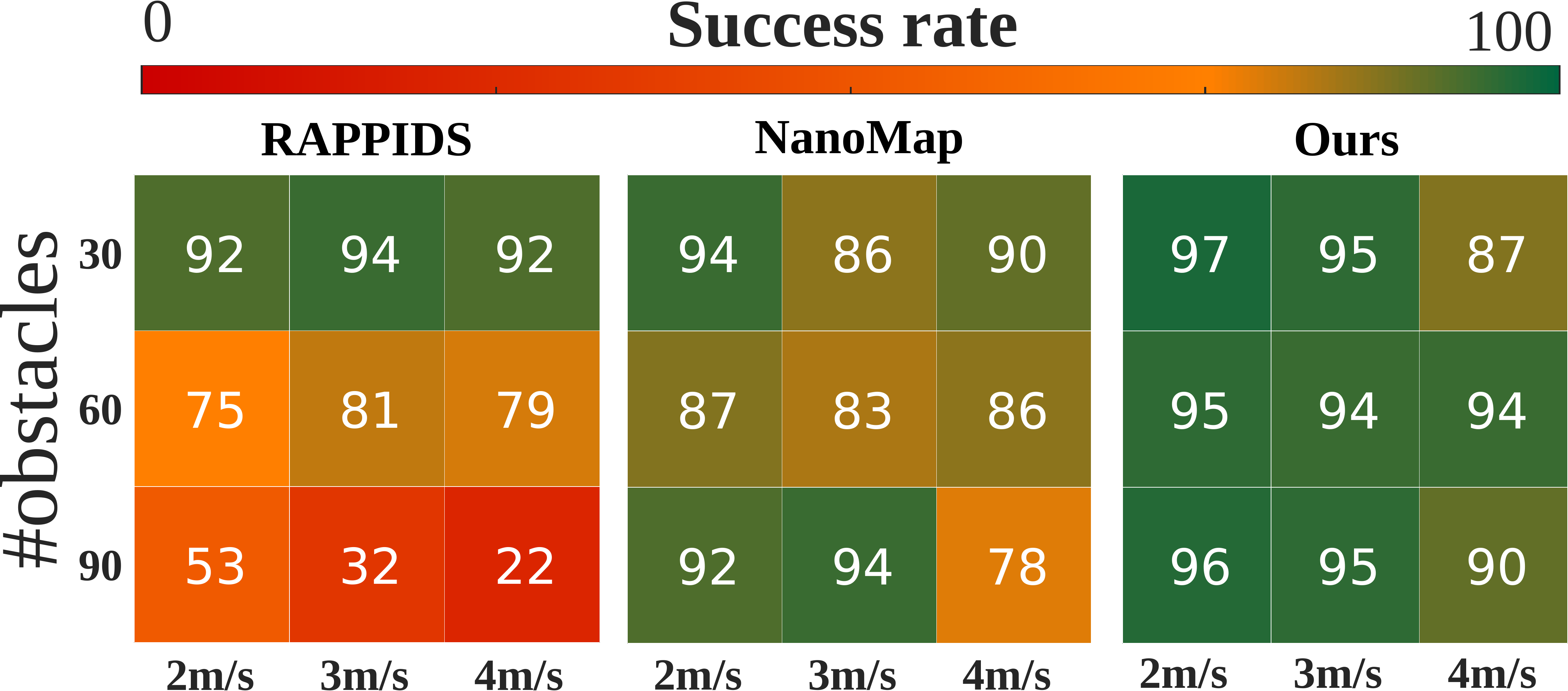}
  \caption{
    \label{fig:success}
    Comparison of number of successful collision-free flights for the different methods tested in our simulated quadrotor race.
  }
\end{figure}

\begin{figure}[ht]
  \centering
  \includegraphics[width=1.0\columnwidth]{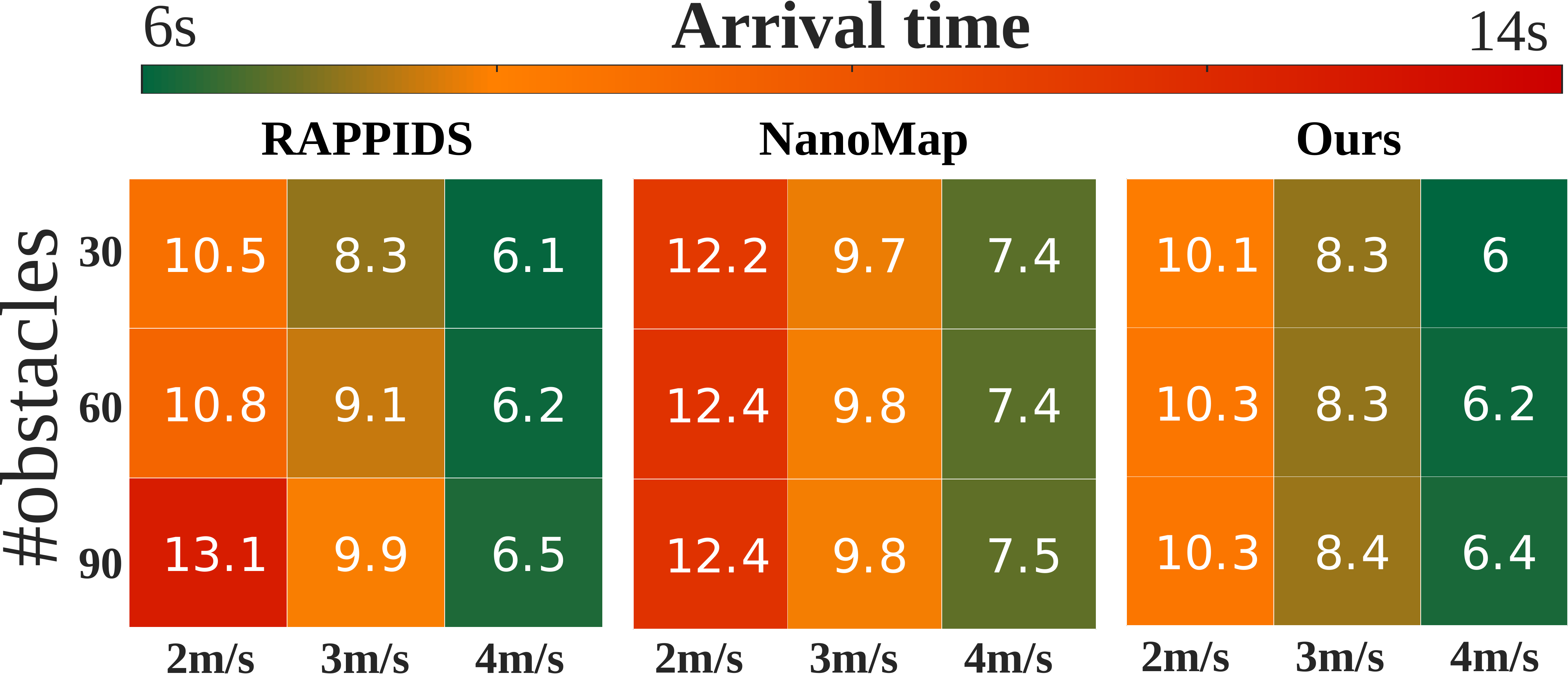}
  \caption{
    \label{fig:time}
    Comparison of the average time to goal for successful flights.
  }
\end{figure}

The key metric for our comparison of these three methods is the collision-free success rate of trials and the average time to reach the goal in successful flights, which are presented in Fig.\ref{fig:success} and Fig.\ref{fig:time}. The results for RAPPIDS show its limitations on handling cluttered environments. For NanoMap, the average time is longer than our method by approximately $19\%$. As is mentioned in Section III, NanoMap may falsely evaluate $\mathbf d_{obstacle}$ ignoring the blind spot. Such a flaw tends to be unsafe and leads a relatively low success rate. These results demonstrate the robustness and aggressiveness of our mapless planner in cluttered environments.

\section{Conclusion and future work}
\label{sec:conclusion}

In this paper, we propose a robust and fast mapless planning framework for aggressive autonomous flight without map fusion, which is able to abstract and exploit the environment information efficiently for generating high-quality trajectories. Benchmark comparisons and real-world experiments validate that it is lightweight, robust, and highly efficient. In the future, we will improve our framework to fit it in smaller quadrotors with coarse localization.

\addtolength{\textheight}{-13.6cm} 

\bibliography{ICRA2020jjl}
\end{document}